\pdfoutput=1

\documentclass[11pt]{article}
\usepackage{fancybox}
\usepackage[]{acl}
\usepackage{times}
\usepackage{latexsym}
\usepackage{ragged2e}

\usepackage[T1]{fontenc}

\usepackage[utf8]{inputenc}

\usepackage{microtype}

\usepackage{adjustbox}
\usepackage{booktabs}
\usepackage{textcomp}
\usepackage{xcolor}
\usepackage{listings}
\usepackage{graphicx}
\usepackage{subcaption}
\usepackage{mwe}
\usepackage{tikz}
\usepackage{pgfplots}
\usetikzlibrary{pgfplots.groupplots}
\usepackage{amssymb} 
\usetikzlibrary{arrows,automata,calc,shapes, positioning,shadows,shadows.blur,shapes.geometric,decorations.pathmorphing,patterns,matrix}
\usepackage{pgfplotstable}
\usepgfplotslibrary{groupplots}
\definecolor{c1}{HTML}{ef476f}
\definecolor{c2}{HTML}{f78c6b}
\definecolor{c3}{HTML}{ffd166}
\definecolor{c4}{HTML}{06d6a0}
\definecolor{c5}{HTML}{118ab2}
\definecolor{c6}{HTML}{073b4c}

\definecolor{codegreen}{rgb}{0,0.6,0}
\definecolor{codegray}{rgb}{0.5,0.5,0.5}
\definecolor{codepurple}{rgb}{0.58,0,0.82}
\definecolor{backcolour}{rgb}{0.9,0.98,0.95}

\newcommand{\draftonly}[1]{#1}
\renewcommand{\draftonly}[1]{}

\usepackage{pifont}
\newcommand{\xmark}{\ding{55}}%

\usepackage{arydshln}  

\newcommand{\presto}[0]{\textsc{PRESTO}}

\title{\presto{}: \textbf{A Multilingual Dataset for \\
Parsing Realistic Task-Oriented Dialogs} \\ \quad}

\author{
\parbox{1.1\linewidth}{\centering
Rahul Goel\thanks{\quad Equal contribution.}~,~~
Waleed Ammar\footnotemark[1]~,~~
Aditya Gupta\footnotemark[1]~,~~
Siddharth Vashishtha\footnotemark[1]~\hspace{0.5mm}\thanks{\quad Work done during an internship at Google.}\hspace{1.2mm}$^{ \clubsuit}$,~~\\
Motoki Sano\footnotemark[1]~,~
Faiz Surani\footnotemark[1]~\hspace{0.5mm}\footnotemark[2]\hspace{1.2mm}$^\diamondsuit$,~
Max Chang,~
HyunJeong Choe,~
David Greene,~
Kyle He,~\\
Rattima Nitisaroj,~
Anna Trukhina,~
Shachi Paul,~
Pararth Shah,~
Rushin Shah,~
Zhou Yu$^\spadesuit$
}
\\ Google Inc.\\ $^\clubsuit$ University of Rochester~~ $^\diamondsuit$ University of  California, Santa Barbara ~~ $^\spadesuit$Columbia University
\\ \texttt{presto.dataset@google.com} 
}

\date{}

\begin{document}
\maketitle

\begin{abstract}
Research interest in task-oriented dialogs has increased as systems such as Google Assistant, Alexa and Siri have become ubiquitous in everyday life.
However, the impact of academic research in this area has been limited by the lack of datasets that realistically capture the wide array of user pain points.
To enable research on some of the more challenging aspects of parsing realistic conversations, we introduce \presto\footnote{\url{https://github.com/google-research-datasets/presto}}, a public dataset of over 550K contextual multilingual conversations between humans and virtual assistants. \presto{} contains a diverse array of challenges that occur in real-world NLU tasks such as disfluencies, code-switching, and revisions. It is the only large-scale human generated conversational parsing dataset that provides structured context such as a user's contacts and lists for each example. 
Our mT5 model based baselines demonstrate that the conversational  phenomenon present in \presto{} are challenging to model, which is further pronounced in a low-resource setup.    
\end{abstract}

\section{Introduction}
Virtual dialog agents (a.k.a. ``\emph{assistants}'') are increasingly becoming a part of our everyday life.
From setting up alarms to controlling one's home environment, users now converse with their devices to accomplish many tasks previously done manually.
Parsing task-oriented dialogs is the problem of understanding the user's intent and any associated arguments so that the assistant can fulfill the requested task.

Given the prominence of data-driven methods in NLP, public availability of relevant datasets dictates which problems can be effectively studied by the research community at large.
Datasets such as MultiWoz~\cite{multiwoz,eric2019multiwoz}, TOP~\cite{top}, MTOP~\cite{mtop}, and SMCalFlow~\cite{smcalflow} have enabled researchers to study numerous developments in conversational semantic parsing, e.g.,~\newcite{Budzianowski2019HelloIG}, ~\newcite{Pasupat2021ControllableSP}.
However, many challenging aspects of parsing, such as disfluencies, code-switching, and the use of the structured context which grounds conversations, have been missing in these datasets, stifling meaningful research progress in those areas.

\begin{figure*}[t]
    \centering
    \includegraphics[width=\textwidth]{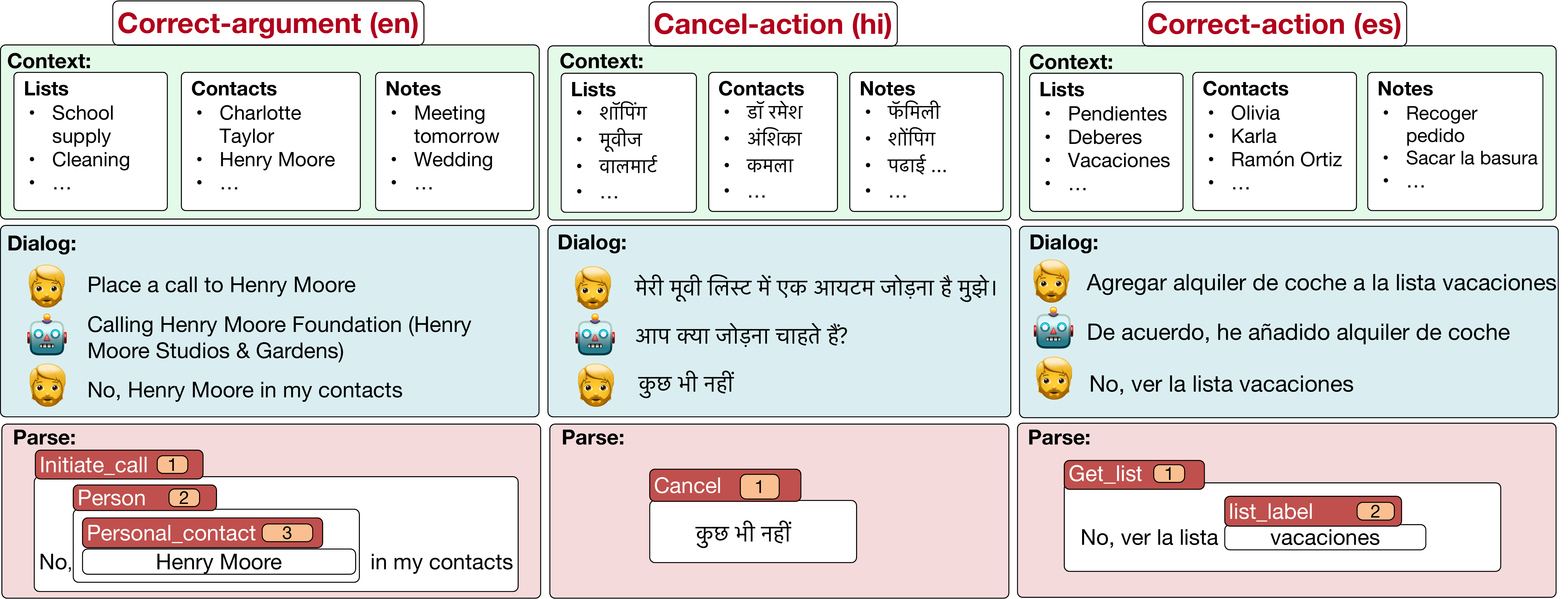}
    \caption{Examples of user revision dialog sessions from \presto. \presto{} includes annotated dialogs in 6 languages (\emph{de, en, es, fr, hi, ja}) with various characteristics such as corrections (\texttt{correct-argument, correct-action}), cancellations (\texttt{cancel-action}) etc. Each example consists of 
\underline{Input}: a user's virtual state (context), one or more user utterances and the corresponding virtual assistant responses (dialog).
\underline{Output}: the semantic parse of the last user utterance in the dialog (parse).}
    \label{fig:example}
\end{figure*}

With that in mind, we present \presto{}, a dataset that better represents real conversations users have with their virtual assistants. \presto{} consists of:
\begin{itemize}
    \item Challenging conversational phenomena such as disfluencies, code-switching and user revisions, addressing some of the aforementioned limitations of current virtual assistants and enabling researchers to test new methods of addressing these challenges. 
    \item Realistic multilingual dialogues. Unlike in other datasets, conversations in each language were not obtained by translating English dialog, but instead contributed by native speakers of each language.
    \item Human and synthetic structured context with the conversations. The data contributors were instructed to look at and optionally reference relevant context in the user's virtual environment. For example, contributors could reference the user's contact list while initiating a phone call or sending a message using the virtual assistant.
\end{itemize}

When users speak naturally to their virtual assistants, we observe a variety of conversational phenomena: disfluent turns~\cite{gupta2021disfl}, user revisions (also known as conversational repair~\cite{cassell2000human}) and code-switching (also known as code-mixing~\cite{agarwal2022cst5}), to name a few.
Our dataset highlights several such linguistic phenomena, and provides over thousands (Table~\ref{tab:data_stats}) of examples of each phenomenon per language, each produced by a native speaker.
Overall, our dataset contains more than 550k annotated utterances across six languages.
Each user utterance is further verified by three native speakers for fluency and correctness and then annotated separately twice. Section \S\ref{sec:data_collection} discusses how the data was collected.

Section \S\ref{sec:data_summary} discusses context as well as other important phenomena which are well-represented in this dataset such as user revisions and disfluent utterances. In section \S\ref{sec:experiments}, we use mT5-based models \citep{Xue2020mT5AM} to present some baseline model performance metrics for this dataset, e.g., exact match accuracy on the test sets, the relative performance of monolingually- vs. multilingually-trained models, and data scaling effects with respect to various linguistic phenomena.

\section{Dataset Characteristics}\label{sec:data_summary}
In this section, we discuss the characteristics of \presto{} dataset that set it apart from existing datasets: native conversations in multiple languages, code-switched utterances, user revisions, disfluencies, and structured context.

\paragraph{Native Speakers.}
The \presto{} dataset only includes utterances provided by native speakers of the language, with no translation. 
Table \ref{tab:data_stats} shows the number of examples in each language.

\begin{table*}[]
\small
\begin{tabular}{lcccccccc}
\toprule
Language & \# Intents / & Avg. Slots / & Avg. Tokens / & Avg. Prev.  & Total & Code & User & Spoken\\
 & \# Slots  & Utterance & Utterance & Utterances & Examples & Switching & Revisions & Disfluencies \\

\midrule
German   & 34 / 285 & 1.60 & 9.17 & 2.56 & 83,584             & 12,357                 & 22,129         & 16,781       \\
English & 34 / 303 &  1.57 & 9.03 & 2.49 & 95,671             & 5,918                  & 27,741         & 17,588       \\
Spanish & 34 / 299 & 1.72 & 10.70 & 2.57 & 96,164            & 12,570                 & 27,713         & 21,510       \\
French  & 34 / 303 & 1.63 & 10.73  & 2.63 & 95,870            & 12,939                 & 25,157         & 20,137       \\
Hindi   & 34 / 285 & 1.47 & 9.32 & 2.55 &72,107            & 16,517                 & 15,833        & 10,193       \\
Japanese & 34 / 292 & 1.73 & 11.23 & 2.58 &	109,528           & 15,200                 & 29,474         & 23,838       \\
\midrule
Total & -  - & 1.63  & 10.11 & 2.57 & 552,924           & 75,501                 & 148,047        & 110,047     \\
\bottomrule
\end{tabular}
\caption{Corpus and sentence level statistics of \presto, slices per language, including various linguistic phenomenon.  }
\label{tab:data_stats}
\end{table*}

Prior large multilingual datasets for conversational semantic parsing such as MTOP~\cite{mtop} and MASSIVE~\cite{massive} contain non-English conversations obtained by translating English conversations to other languages,\footnote{The translation was constrained in order to ensure the English source and the translation have the same semantic parse with argument values in English translated to the corresponding phrases in the other language.} resulting in \textit{unnatural} and synthetic utterances which are unlikely to be spoken by native speakers of the non-English language.
For example, an Arabic example in MASSIVE:\\
\vspace{-2em}
\begin{figure}[h]
\centering
\includegraphics[width=7.5cm]{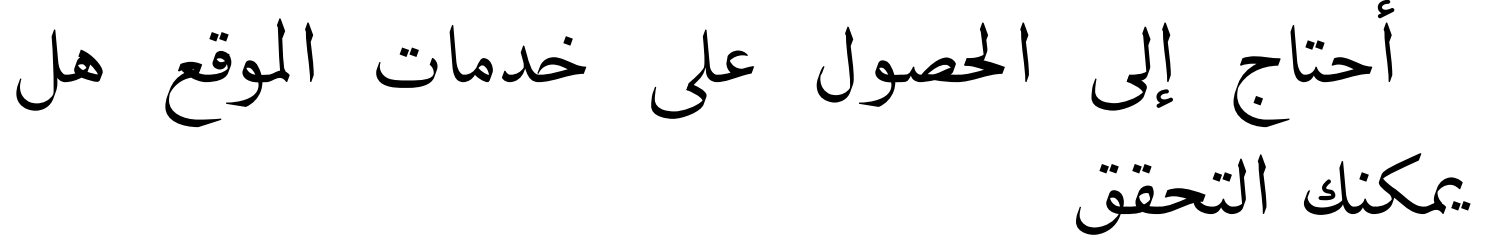}
\end{figure}

\vspace{-1em}
\noindent does not resemble native Arabic speech of any dialect. The corresponding English utterance in the MASSIVE dataset is: \textit{I need to have location services on can you check}

\begin{figure}[h]
\centering
\includegraphics[width=7.5cm]{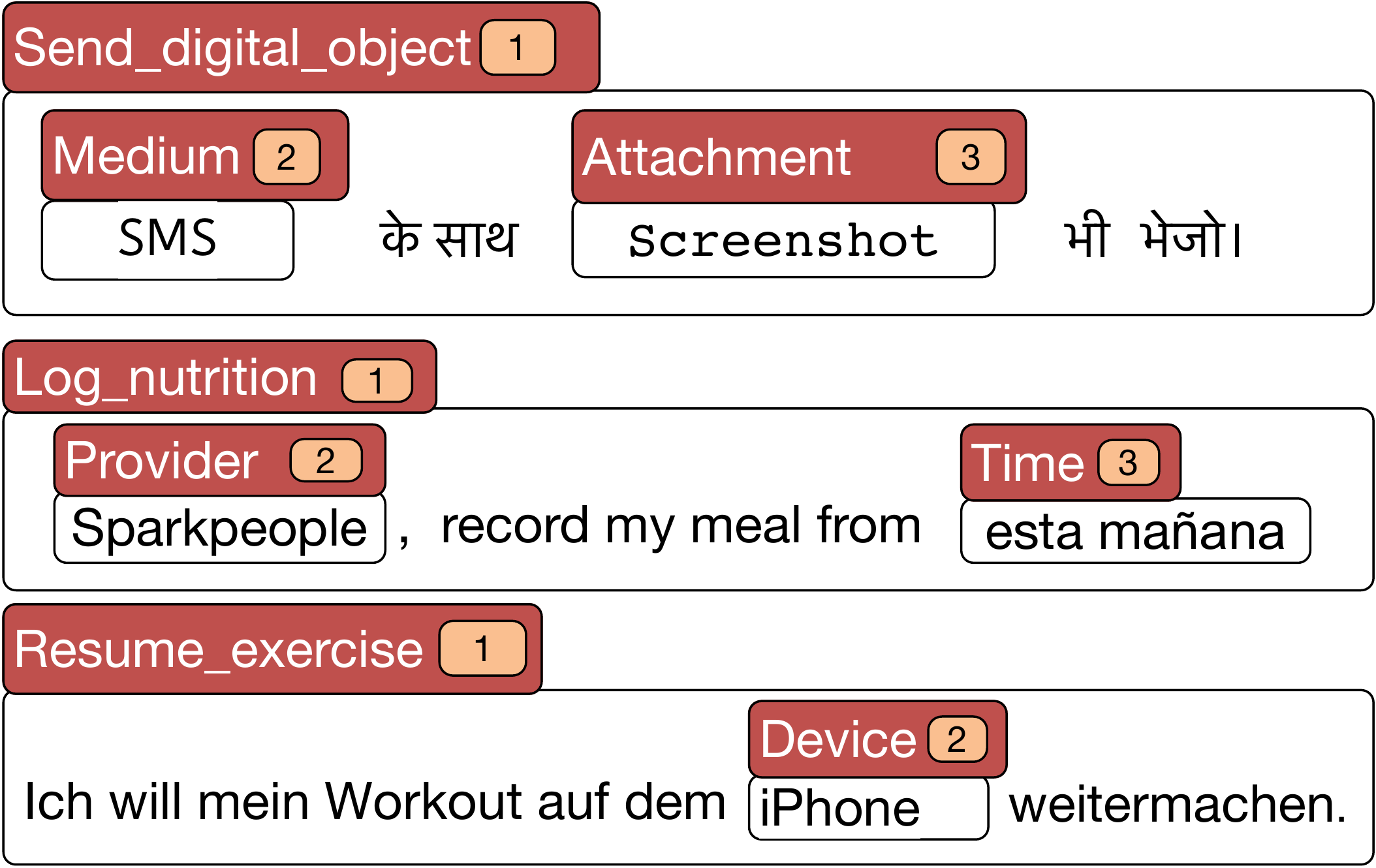}
\caption{Examples of Hindi-English (\textit{Send a screenshot along with the SMS.}), Spanish-English (\textit{Sparkpeople, record my meal from this morning.}), and German-English (\textit{I want to continue my workout on the IPhone.}) code-switched utterances from \presto{}.}
\label{fig:codeswitching}
\end{figure}
\paragraph{Code-Switched Utterances.}
Multilingual users often mix words from two languages in the same utterance, as shown in Fig.~\ref{fig:codeswitching}.
Recognizing the difficulty of parsing such utterances, we asked bilingual data contributors to provide mixed utterances, and marked these examples with a special tag `\texttt{code-switching}' to enable isolated research focused on addressing this phenomenon. 
Table \ref{tab:data_stats} shows the number of code-switched examples in each language.\footnote{We use a language ID classifier to determine which of the two languages is dominant in a given utterance.}

\paragraph{User Revisions.} 

The example in Fig.~\ref{fig:example} highlights another important aspect of realistic conversations: users often revise their request to the assistant. We have 4 such tags in our dataset which are all grouped under the broader user-revision linguistic phenomenon. The tags are: \texttt{correct-action}, \texttt{correct-argument}, \texttt{within-turn-correction}, and \texttt{cancel- action}.  
Sometimes the revision is necessary due to a mistake made by the virtual assistant, as in Fig.~\ref{fig:example} (\textit{correct-argument}). At other times, the user may simply change their mind about what they want the assistant to do which may happen in the same utterance, e.g., ``\textit{Add bread no no wait add wheat bread to my shopping list}'', or in a later utterance, e.g., ``\textit{Sorry add wheat bread instead}.''
Another common revision users make is to cancel their requests, e.g., ``\textit{Sorry don't add anything}''. Fig.~\ref{fig:example} shows examples of some of these revisions. Table~\ref{tab:data_stats} reflects the number of examples with user revisions in the last utterance for each language in \presto{}.

\begin{figure}[h]
\centering

\includegraphics[width=7.5cm]{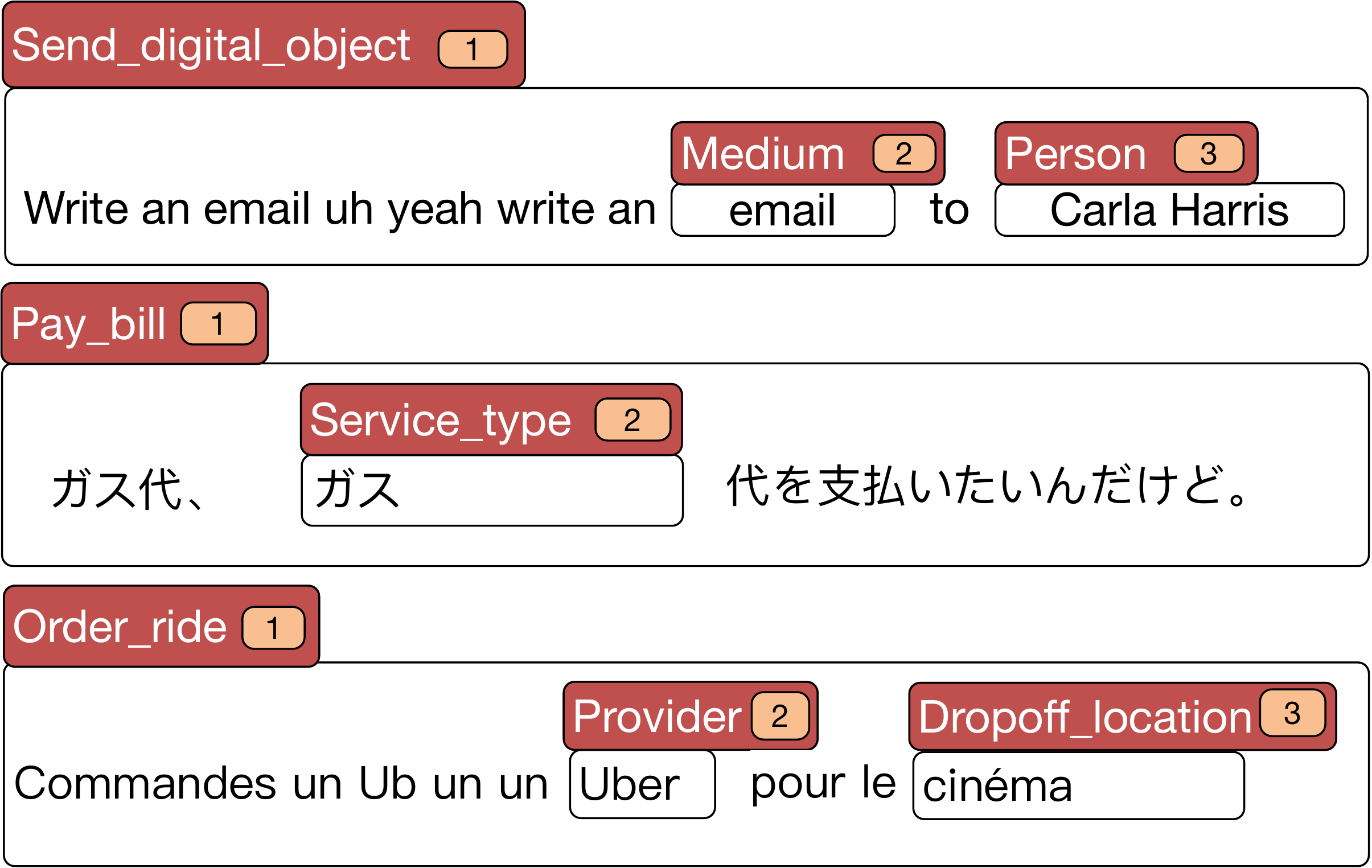}
\caption{Examples of utterances in English, Japanese (\textit{Gas bill, I want to pay the gas bill.}), and French (\textit{Order an Ub an Uber for the cinema}) with disfluencies (filler words or repetitions) from \presto{}.}
\label{fig:disfluency}
\end{figure}

\paragraph{Disfluencies.}
Due to the spoken nature of most conversations with virtual assistants, user utterances frequently have disfluencies such as repeated phrases and filler words, as shown in Fig.~\ref{fig:disfluency}. Table~\ref{tab:data_stats} shows the number of examples with disfluencies for each language in \presto{}.

\paragraph{Structured Context.}
A typical user interacts with virtual assistants within a virtual world (henceforth \emph{context}) which may consist of various structured objects, e.g., a list of contacts on the user's phone, a shopping list, a to-do list, notes created by the user and others sent from family and friends, a recurrent 5:45 am alarm to get out of bed every weekday, 4 smart bulbs, 12 smart plugs, a smart sprinkler in the backyard, etc.

Depending on the query, this context may or may not be needed to correctly interpret user utterances; it is often hard for semantic parsing models to determine which part of the context matters for a given utterance, if any. Moreover, the modeling solutions should have the ability to model (and ignore) such structured information. Fig.~\ref{fig:example} (\texttt{correct-argument}) shows an example in our dataset where the virtual assistant fails to relate the `\textit{Henry Moore}' to the user's contacts in the structured contact and incorrectly interprets the name as a reference to the ``\textit{Henry Moore Foundation}''. 
The gold parse provided in the dataset shows the correct interpretation for the last utterance, leveraging the structured context.

\begin{table*}[]
\resizebox{\textwidth}{!}{\begin{tabular}{lcccc}
\toprule
Dataset & \# Languages & Multi-turn & Explicit  & Labeled Conversational  \\
 & & & Context & Phenomena \\
\midrule
\textbf{\presto{}} & 6 &  \checkmark & \checkmark & Code-Switching, Disfluencies, \\
 & & & & User-Revisions, Coreferences$^*$\\
 \midrule
TreeDST \citep{treedst} & 1 & \checkmark & \xmark & Coreferences \\
SMCalFlow \citep{smcalflow} & 1 & \checkmark & \xmark & Coreferences, User-Revisions \\
MultiWOZ \citep{multiwoz}& 1 & \checkmark &  \xmark & Coreferences$^*$\\
DSTC10 \citep{dstc10} & 1 & \checkmark & \xmark & Disfluencies$^*$ \\
TOP \citep{top}& 1 & \xmark  & \xmark & - \\
MTOP \citep{mtop}& 6 & \xmark & \xmark & - \\
MASSIVE \citep{massive}& 51 & \xmark & \xmark & - \\
SNIPS \citep{snips} & 1 & \xmark & \xmark & -\\
MultiATIS++ \citep{multiatis} & 9 & \xmark & \xmark & -\\
\bottomrule
\end{tabular}
}
\caption{ \small Comparison of \presto{} with other recent datasets on or related to semantic parsing of task-oriented dialog. * denotes phenomena which is implicitly present in the dataset but not explicitly tagged.
}
\label{tab:data_comparison}
\end{table*}

\paragraph{Other Statistics.}
\presto{} covers 34 intents, including 
the intent `\texttt{Other}' which is a catch-all intent used for all out-of-scope utterances and constitutes 11.3\% of all examples in the dataset.
The average number of the previous user turns is around 2.6, which is representative of the relatively short conversations users tend to have with virtual assistants today.
Table \ref{tab:data_stats} provides summary statistics for the number of tokens in labeled user utterances in each language, demonstrating the large variety in utterance lengths.

\paragraph{Related Datasets}
There have been multiple related datasets which are listed in Table \ref{tab:data_comparison}. The existing task-oriented dialog datasets can be split by capabilities and representations. The salient capabilities are -- single turn or conversational, monolingual or multilingual, presence of conversational phenomenon, and  single-domain vs multi-domain. In terms of representations, there are 2 popular paradigms, a direct semantic parse  typically consisting of intents and slots vs maintaining an explicit dialog state. With the popularity of pretrained text-to-text models, the natural modeling choice for all of these has converged to similar models. In fact, in terms of models, our context representation can be seen as a constant dialog state for that conversation.

\section{Data Collection}\label{sec:data_collection}
The scope for data collection is based on the App Actions built-in intents, by Google, available for 3rd-party Android Developers, which cover a wide spectrum of domains including finance, health, fitness, food, transportation, social, communications, shopping among others.\footnote{\url{https://developer.android.com/reference/app-actions/built-in-intents}} Our dataset has over 34 unique intents and 370 arguments across 6 languages. Table~\ref{tab:intent_distribution} talks about the intent distribution in the dataset.  
We use two complementary approaches to collect contextual and non-contextual examples.

\paragraph{Contextual Examples.}
Data contributors were asked to have conversations with a virtual assistant simulator based on different sets of instructions for each language and for each type of targeted linguistic phenomenon (e.g., \texttt{code-switching}).
Each data collection request targeted a single intent and was initialized with potentially relevant structured context, e.g., contacts, notes or lists, which can optionally be referenced by data contributors in the utterances they produce.
The initial seed contacts, notes and lists are authored by native speakers and are shared for all examples in one request.
For example, when initiating a call, data contributors may use one of the names in the contacts (e.g., ``\textit{call mom}''), but they may also use the name of a business not in the contact list (e.g., ``\textit{call mcdonalds}'').

Each user utterance from the collected conversations is exported as a candidate example for semantic annotation by trained expert linguists who produce two types of annotations.\footnote{Due to limited annotation capacity, we were not able to annotate all candidate examples and we prioritized candidate examples which have at least one previous turn. Candidates which were not included were discarded and are not included in this  data release.}
First, linguists decide whether the example is in scope or not.
Out of scope examples include incoherent or nonsensical utterances, unsupported arguments and out-of-scope intents.
Then, they choose the intent and nested arguments as shown in Fig.~\ref{fig:example}. 
Finally, they added semantic tags which indicate which linguistic phenomena of interest are expressed in this example, such as, \texttt{correct-argument}, \texttt{cancel-action}, \texttt{code-switching}, etc.

\paragraph{Non-Contextual Examples.}
A key challenge in representing context in real world semantic parsing is that most examples can be parsed without using it, for example utterances like ``\textit{play music}'' are typically non-contextual but an utterance like ``\textit{play workout music}'' might be contextual if the user has a playlist called ``\textit{workout}''. This makes it a challenging modeling task; context ideally should only be used when it is relevant. On the one hand, exclusively using relevant contextual examples when training a conversational semantic parser results in models which are over-sensitive to irrelevant contextual features.
On the other hand, exclusively using non-contextual training examples where the user utterance is not related to previous turns or structured context teaches the model that the context does not matter, and may result in emphasizing arbitrary contextual features which happen to correlate with some prediction in the training data. 

A balanced approach is to include both types of examples in training, and use targeted evaluation sets to better understand the impact of different ways of representing context.
We obtain non-contextual examples by asking linguists to produce single-turn utterances along with their gold parses, then pair them with samples of the structured context and previous turns from the contextual conversations discussed earlier to make the representation uniform. The pairing method is described in the post-processing section. We denote such synthetic generated context in the data by the tag \texttt{context: synthetic}. 

\paragraph{Data Quality.} 
We adopt several mechanisms for boosting the data quality of collected data. All our data collection is done in 2 phases. Phase 1 is data collection, where annotators come up with queries pertaining to a scenario. Phase 2 is the actual semantic parse annotation for the query. 

For non-contextual examples, we use a crowd compute like platform with native speakers to generate and annotate the data. The process is described below:
\begin{enumerate} 
    \item \texttt{\textbf{[Phase 1.1]}} Annotator 1 authors a query (contextual or non-contextual). 
    \item \texttt{\textbf{[Phase 1.2]}} Annotators 2, 3, 4 independently validate the query (scope judgment). If all 2, 3, 4 agree that the query is valid, then the query goes to phase 2. Upon any disagreement, the example is discarded.
    \item \texttt{\textbf{[Phase 2.1]}} Annotators 5 and 6 annotate the query independently. If their annotations match, the query is considered resolved along with the annotation.  
    \item \texttt{\textbf{[Phase 2.2]}}  If 5, 6 disagree, the query is sent to annotator 7 who annotates the query. If 7 matches any of 5 or 6, then the query is resolved to the matching annotation else the example is discarded. 
\end{enumerate}
As can be seen, the overall query generation and annotation process is similar for contextual and non-contextual examples. However, contextual examples require more care in annotation therefore we additionally ask trained in-house linguists who also speak the language to verify the plausibility of each example and annotations and exclude ones that are out of scope or unclear as discussed earlier.

In order to verify the efficacy of our approach, we manually analyze a random sample of $6K$ examples and evaluate the correctness of the parse and the plausibility of the user utterance.
The three labels available for assessing the semantic parse quality are: `\texttt{Accuracy:GOOD}', `\texttt{Accuracy:BAD:INTENT}', `\texttt{Accuracy:BAD:SLOT}'.
The two labels available for assessing the utterance are: `\texttt{Acceptability:HIGH}' and `\texttt{Acceptability:LOW}'.

We use the \texttt{Accuracy:GOOD} label when the top-level intent and all slot values are accurate according to  the annotation guidelines.
\texttt{Accuracy:BAD:INTENT} is used when an intent is not accurate based on the annotation guidelines, while \texttt{Accuracy:BAD:SLOT} is used when the intent is accurate but one or more slot values are not accurate.
We use the \texttt{Acceptability:HIGH} label when the utterance is likely to be spoken or typed by Assistant users, and is grammatical and contextually correct, otherwise we use \texttt{Acceptability:BAD}.
Before post-processing, we observe that the annotators agree on utterance plausibility (acceptability) for 92.9\% of  examples,
and agree on full semantic parse annotation (accuracy) for 90.7\% of examples.  

The breakdown of intent and argument errors in contextual examples can be found in Table~\ref{tab:errors} for five of the six languages (Spanish was not included in this analysis due to capacity constraints). 

\begin{table}[h]
\centering
\begin{tabular}{lcc} 
\toprule
Language & Bad Intent & Bad Argument \\\midrule
German & 2.8\% & 3.7\% \\
English & 2.9\% & 8.5\% \\
French & 0.0\% & 2.5\% \\
Hindi & 1.9\% & 4.8\% \\
Japanese & 4.3\% & 2.1\%
\\\bottomrule
\end{tabular}
\caption{Type of accuracy errors in  examples by language from our human evaluation.}
\label{tab:errors}
\end{table}

\paragraph{Post-processing.}
One of the common annotation errors we found is that code-switched examples often had the wrong language ID (\texttt{LangID}) associated with them. To address this, we use a language ID classifier to determine the dominant language in code-switched examples.

Some arguments (due to fine-grained domain modeling) were often confused for other arguments which have similar (but distinct) semantics, e.g., \texttt{ListItem} vs. \texttt{ExistingListItem}, resulting in inconsistent annotations and contributing to the error rate discussed earlier.
To address this, we merged pairs of argument names where the distinction rarely matters in practice.

To make our data uniform, we add synthetic context and previous turns to non-contextual examples. Examples which are missing previous turns are augmented to reuse previous turns from other examples by first sampling the number of previous turns using a negative binomial distribution with $r=5, p=0.8$, then sampling one of the collected conversations which has this many turns. 
Examples which are missing a given type of structured context, e.g., contacts, are augmented by randomly sampling contact names from related argument values specified by our data contributors, which results in realistic and diverse contexts.

We do another quality check to measure Accuracy and Acceptability after post-processing by randomly sampling 500 queries from each language (except Spanish due to capacity constraints). We ask 3 independent annotators (same as the earlier quality check) to rate the queries. After post-processing, we find that linguists agree, defined by 2 out of 3 annotators independently agreeing, on utterance plausibility (acceptability) for \textbf{98.8\%} of  examples and agree on full semantic parse annotation (accuracy) for \textbf{93.24\%} of examples.  This shows our post-processing decreased the data noise significantly.

\section{Experiments}\label{sec:experiments}
In this section, we demonstrate a few experimental setups which are enabled by \presto{}, provide baseline results for future work, and summarize our findings.

\subsection{Setup}\label{sec:setup}
Each example in the dataset is designated as train, development or test with respective probabilities 0.50, 0.15 and 0.35, which enables us to have large test sets even when doing focused evaluation on a particular phenomenon of interest as in \S\ref{sec:experiments_code_switching}.
All data splits are provided as part of the data release.
We use the \texttt{t5x} library \cite{Roberts2022ScalingUM} to fine-tune mT5's public checkpoint \cite{Xue2020mT5AM} on the train portion of the data (unless otherwise stated in k-shot experiments).

For all experiments, we report \textbf{exact match accuracy} of the predicted semantic parse, which gives the model one point for each example where the intent and all the arguments are correctly predicted and zero points otherwise.
All experiments are based on the mT5-Base model (580M parameters),\footnote{\url{https://github.com/google-research/t5x/blob/main/t5x/examples/t5/mt5/base.gin}}
except the scaling experiments in \S\ref{sec:experiments_model_scaling} which demonstrate the effect of model scaling to mT5-Large (1.2B parameters), mT5-XL (3.7B parameters), and mT5-XXL
(13B parameters).
All models (except the monolingual models discussed in \S\ref{sec:experiments_monolingual}) are fine-tuned on the union of training examples from all languages in \presto{}.
Few shot experiments for code switching (\S\ref{sec:experiments_code_switching}), user revisions (\S\ref{sec:experiments_revisions}) and disfluencies (\S\ref{sec:experiments_disfluencies}) all share the same training set, e.g., the 5-shot model across all three sections consist of 5 code-switching examples, 5 cancellations, 5 within-turn corrections, 5 intent cross-turn corrections, 5 argument cross-turn corrections and 5 disfluency examples, in addition to all examples which do not represent any of these phenomena.

\paragraph{Features.} 
\label{para:features}
The input sequence fed to the model consists of the last user utterance which needs to be parsed, followed by previous turns (both user and assistant turns) in reverse chronological order, followed by some representation of the structured context (more on this in \S\ref{sec:context}), with a separator token between consecutive fields. An example is shown in Figure~\ref{fig:data_serialization}.

\begin{figure}[]
   {\small \textbf{Input:} \\ 
    \texttt{No, Henry Moore in my contacts | Calling Henry Moore Foundation (Henry Moore Studios \& Gardens) | Place a call to Henry Moore [SEP] Lists: School Supply, Cleaning, ... [SEP] Contacts: Charlotte Taylor, Henry Moore, ... [SEP] Notes: Meeting tomorrow, ...}
    \\
    \\
    {\small \textbf{Output:}\\
    \texttt{Initiate\_call( callee = Personal\_contact( person = Henry Moore ) ) }
    }%
}

    \caption{Example serialization and the output of the features in the dataset. The top box shows the encoding for the first input in Fig.~\ref{fig:example}. The output sequence consists of a depth-first traversal of the semantic parse, is also shown.  }
    \label{fig:data_serialization}
\end{figure}

\paragraph{Hyper-parameters.}
We do a rough hyperparameter tuning and fine-tune the model for 20K steps. We use the Adafactor optimizer with 0.8 decay rate and 0 step offset. We have a max input length of 512 for training and 1024 for inference and a batch size of 128. 

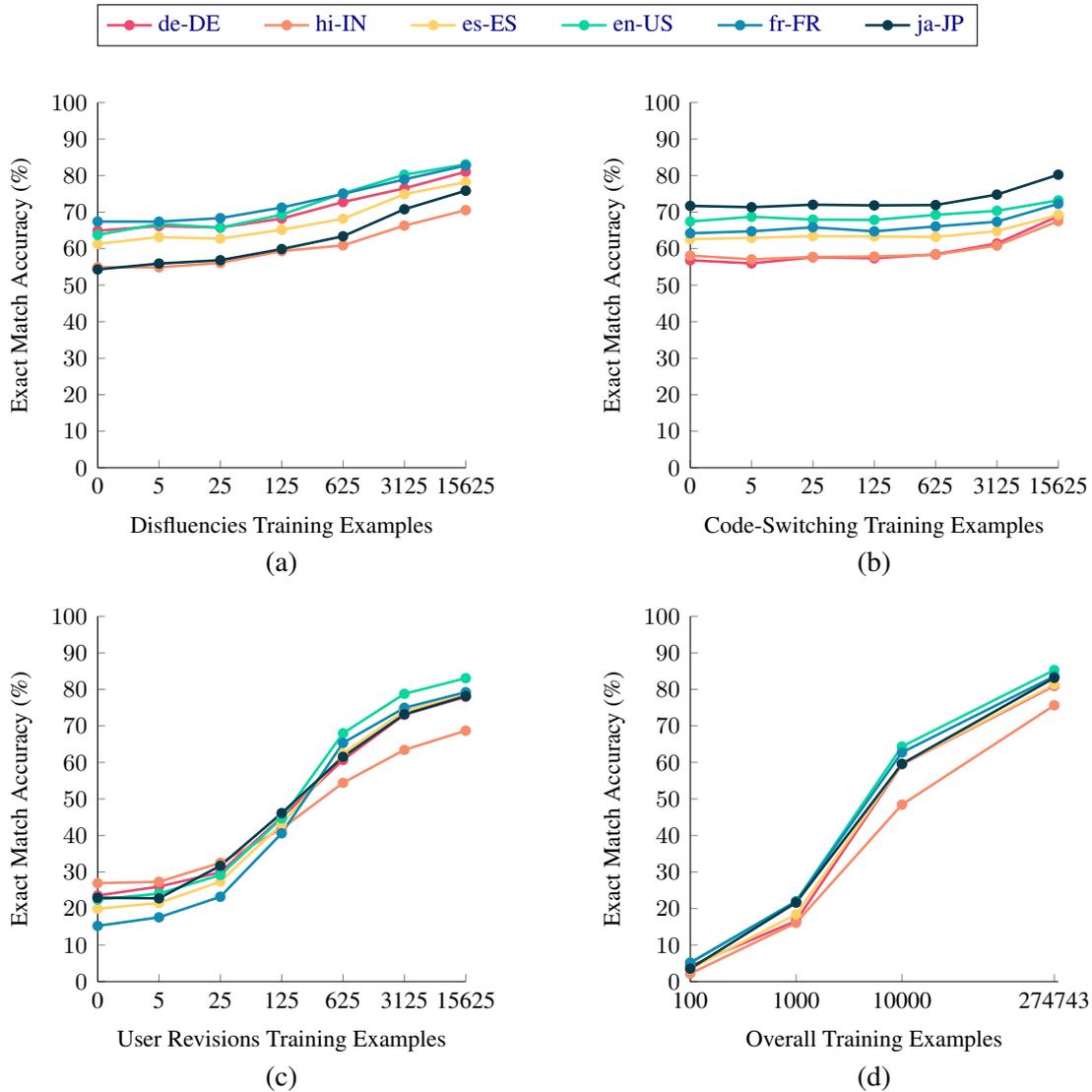
\begin{figure*}[t]
  \begin{centering}
    \begin{tikzpicture}
      \pgfplotsset{footnotesize,samples=10}
      \begin{groupplot}[group style = {group name=my plots, group size = 2 by 2, horizontal sep = 3cm, vertical sep = 2cm},
          width = 6.5cm,
          height = 6.5cm]
        \nextgroupplot[
          align = left,
          xmin=1, xmax=15625,
          xmode=log,
          ymin=0, ymax=100,
          axis x line*=top,
          axis y line*=left,
          xlabel={Disfluencies Training Examples},
          ylabel={Exact Match Accuracy (\%)},
          ytick={0, 10, 20, 30, 40, 50, 60, 70, 80, 90, 100},
          xtick={1, 5, 25, 125, 625, 3125, 15625},
          xticklabels={0, 5, 25, 125, 625, 3125, 15625},
          grid style=dashed,
          x label style={at={(axis description cs:0.5,0)},anchor=north},
          y label style={at={(axis description cs:0.1,0.5)},anchor=south},
          xtick pos=bottom,
          ytick pos=left,
          legend style = { /tikz/every even column/.append style={column sep=0.5cm}, legend columns = -1, legend to name = grouplegend,},  
          xmode=log,
        ]
        \addplot[ %
          color=c1,
          mark=*,
          mark size=1.5pt,
          line width=1pt,
        ]
        coordinates {
            (1, 64.91347132677299)
            (5, 66.1689854088904)
            (25, 65.7787580590431)
            (125, 68.2219205972175)
            (625, 72.75195113674924)
            (3125, 76.50152697658635)
            (15625, 81.03155751611808)
          };
        \addlegendentry{de-DE}
        \addplot[ %
          color=c2,
          mark=*,
          mark size=1.5pt,
          line width=1pt,
        ]
        coordinates {
            (1, 54.892667967661005)
            (5, 54.892667967661005)
            (25, 56.09144131586283)
            (125, 59.325341511011985)
            (625, 60.88653470867019)
            (3125, 66.29495400055757)
            (15625, 70.53247839420128)
          };
        \addlegendentry{hi-IN}
        \addplot[ %
          color=c3,
          mark=*,
          mark size=1.5pt,
          line width=1pt,
        ]
        coordinates {
            (1, 61.3082627118644)
            (5, 63.16207627118644)
            (25, 62.725105932203384)
            (125, 65.1615466101695)
            (625, 68.19385593220339)
            (3125, 74.9073093220339)
            (15625, 78.2176906779661)
          };
        \addlegendentry{es-ES}
        \addplot[ %
          color=c4,
          mark=*,
          mark size=1.5pt,
          line width=1pt,
        ]
        coordinates {
            (1, 63.77914507772021)
            (5, 66.75841968911918)
            (25, 65.73834196891191)
            (125, 69.28432642487047)
            (625, 75.14572538860104)
            (3125, 80.22992227979275)
            (15625, 83.06347150259067)
          };
        \addlegendentry{en-US}
        \addplot[ %
          color=c5,
          mark=*,
          mark size=1.5pt,
          line width=1pt,
        ]
        coordinates {
            (1, 67.41652518392756)
            (5, 67.38822863610639)
            (25, 68.35031126202603)
            (125, 71.25070741369552)
            (625, 74.98585172608941)
            (3125, 78.94736842105263)
            (15625, 82.75325410299943)
          };
        \addlegendentry{fr-FR}       
        \addplot[ %
          color=c6,
          mark=*,
          mark size=1.5pt,
          line width=1pt,
        ]
        coordinates {
            (1, 54.30519014589811)
            (5, 55.90767758909352)
            (25, 56.828509925855066)
            (125, 59.87801961253288)
            (625, 63.38196603683329)
            (3125, 70.78450131547477)
            (15625, 75.85505859842144)
          };
        \addlegendentry{ja-JP}
        \nextgroupplot[
          align = center,
          xmin=1, xmax=15625,
          xmode=log,
          ymin=0, ymax=100,
          axis x line*=top,
          axis y line*=left,
          xlabel={Code-Switching Training Examples},
          ylabel={Exact Match Accuracy (\%)},
          ytick={0, 10, 20, 30, 40, 50, 60, 70, 80, 90, 100},
          xtick={1, 5, 25, 125, 625, 3125, 15625},
          xticklabels={0, 5, 25, 125, 625, 3125, 15625},
          grid style=dashed,
          x label style={at={(axis description cs:0.5,0)},anchor=north},
          y label style={at={(axis description cs:0.1,0.5)},anchor=south},
          xtick pos=bottom,
          ytick pos=left,
          xmode=log,
        ]
        \addplot[ %
          color=c1,
          mark=*,
          mark size=1.5pt,
          line width=1pt,
        ]
        coordinates {
            (1, 56.81139444061567)
            (5, 55.96140592694693)
            (25, 57.63841029175282)
            (125, 57.36273834137376)
            (625, 58.44245348035837)
            (3125, 61.40592694693316)
            (15625, 68.89501493223065)
          };
        \addplot[ %
          color=c2,
          mark=*,
          mark size=1.5pt,
          line width=1pt,
        ]
        coordinates {
            (1, 58.097863542384566)
            (5, 57.06409372846313)
            (25, 57.718814610613364)
            (125, 57.83942108890421)
            (625, 58.32184700206754)
            (3125, 60.837353549276365)
            (15625, 67.53962784286699)
          };
        \addplot[ %
          color=c3,
          mark=*,
          mark size=1.5pt,
          line width=1pt,
        ]
        coordinates {
            (1, 62.60180995475113)
            (5, 62.94117647058823)
            (25, 63.393665158371036)
            (125, 63.32579185520362)
            (625, 63.19004524886878)
            (3125, 64.7737556561086)
            (15625, 69.27601809954751)
          };
        \addplot[ %
          color=c4,
          mark=*,
          mark size=1.5pt,
          line width=1pt,
        ]
        coordinates {
            (1, 67.43185078909613)
            (5, 68.72309899569584)
            (25, 67.95791487326638)
            (125, 67.8622668579627)
            (625, 69.2491630798661)
            (3125, 70.34911525585844)
            (15625, 73.21855571496891)
          };
        \addplot[ %
          color=c5,
          mark=*,
          mark size=1.5pt,
          line width=1pt,
        ]
        coordinates {
            (1, 64.2072902942468)
            (5, 64.75625823451911)
            (25, 65.83223539745279)
            (125, 64.73429951690821)
            (625, 66.0737812911726)
            (3125, 67.3693456302152)
            (15625, 72.2880983750549)
          };
        \addplot[ %
          color=c6,
          mark=*,
          mark size=1.5pt,
          line width=1pt,
        ]
        coordinates {
            (1, 71.7065868263473)
            (5, 71.35104790419162)
            (25, 72.0247005988024)
            (125, 71.8375748502994)
            (625, 71.9311377245509)
            (3125, 74.79416167664671)
            (15625, 80.23952095808383)
          };
        \nextgroupplot[
          align = center,
          legend style={at={(0.9,0.05)},anchor=south},
          xmin=1, xmax=15625,
          xmode=log,
          ymin=0, ymax=100,
          axis x line*=top,
          axis y line*=left,
          xlabel={User Revisions Training Examples},
          ylabel={Exact Match Accuracy (\%)},
          ytick={0, 10, 20, 30, 40, 50, 60, 70, 80, 90, 100},
          xtick={1, 5, 25, 125, 625, 3125, 15625},
          xticklabels={0, 5, 25, 125, 625, 3125, 15625},
          grid style=dashed,
          x label style={at={(axis description cs:0.5,0)},anchor=north},
          y label style={at={(axis description cs:0.1,0.5)},anchor=south},
          xtick pos=bottom,
          ytick pos=left,
          legend style={draw=none},
          legend cell align=right,
          xmode=log,
        ]
        \addplot[ %
          color=c1,
          mark=*,
          mark size=1.5pt,
          line width=1pt,
        ]
        coordinates {
            (1, 23.564254442441413)
            (5, 25.997939737316507)
            (25, 29.899562194179758)
            (125, 44.93947978367242)
            (625, 60.66185938707185)
            (3125, 73.0620654133402)
            (15625, 77.96806592840588)
          };
        \addplot[ %
          color=c2,
          mark=*,
          mark size=1.5pt,
          line width=1pt,
        ]
        coordinates {
            (1, 26.936872309899567)
            (5, 27.31348637015782)
            (25, 32.49641319942611)
            (125, 41.85796269727403)
            (625, 54.41176470588235)
            (3125, 63.4325681492109)
            (15625, 68.68723098995696)
          };
        \addplot[ %
          color=c3,
          mark=*,
          mark size=1.5pt,
          line width=1pt,
        ]
        coordinates {
            (1, 19.940415040065748)
            (5, 21.522498459009658)
            (25, 27.35771522498459)
            (125, 43.1066365317444)
            (625, 62.43065543455928)
            (3125, 74.11136223546333)
            (15625, 78.4466817341278)
          };
        \addplot[ %
          color=c4,
          mark=*,
          mark size=1.5pt,
          line width=1pt,
        ]
        coordinates {
            (1, 22.448141302115424)
            (5, 24.08092010679811)
            (25, 29.174368453481208)
            (125, 44.649825426165535)
            (625, 67.98110494968167)
            (3125, 78.78414458821113)
            (15625, 83.05606900800986)
          };
        \addplot[ %
          color=c5,
          mark=*,
          mark size=1.5pt,
          line width=1pt,
        ]
        coordinates {
            (1, 15.25)
            (5, 17.59090909090909)
            (25, 23.21590909090909)
            (125, 40.590909090909086)
            (625, 65.32954545454545)
            (3125, 74.94318181818181)
            (15625, 79.22727272727272)
          };
        \addplot[ %
          color=c6,
          mark=*,
          mark size=1.5pt,
          line width=1pt,
        ]
        coordinates {
            (1, 22.946859903381643)
            (5, 22.792270531400966)
            (25, 31.70048309178744)
            (125, 46.11594202898551)
            (625, 61.51690821256038)
            (3125, 73.17874396135265)
            (15625, 78.16425120772948)
          };
        \nextgroupplot[
          align = center,
          legend style={at={(0.9,0.05)},anchor=south},
          xmin=100, xmax=300000,
          xmode=log,
          ymin=0, ymax=100,
          axis x line*=top,
          axis y line*=left,
          xlabel={Overall Training Examples},
          ylabel={Exact Match Accuracy (\%)},
          ytick={0, 10, 20, 30, 40, 50, 60, 70, 80, 90, 100},
          xtick={100, 1000, 10000, 274743},
          xticklabels={100, 1000, 10000, 274743},
          grid style=dashed,
          x label style={at={(axis description cs:0.5,0)},anchor=north},
          y label style={at={(axis description cs:0.1,0.5)},anchor=south},
          xtick pos=bottom,
          ytick pos=left,
          legend style={draw=none},
          legend cell align=right,
          xmode=log,
        ]
        \addplot[ %
          color=c1,
          mark=*,
          mark size=1.5pt,
          line width=1pt,
        ]
        coordinates {
            (100, 4.259379152894674)
            (1000, 16.625208709578494)
            (10000, 59.46774798105429)
            (274743, 81.00998398473439)
          };
        \addplot[ %
          color=c2,
          mark=*,
          mark size=1.5pt,
          line width=1pt,
        ]
        coordinates {
            (100, 2.1901266721913104)
            (1000, 16.07671362613946)
            (10000, 48.44323428436132)
            (274743, 75.63237441300659)
          };
        \addplot[ %
          color=c3,
          mark=*,
          mark size=1.5pt,
          line width=1pt,
        ]
        coordinates {
            (100, 3.4365279217893647)
            (1000, 18.391349429714115)
            (10000, 59.62375944304548)
            (274743, 81.31536068730558)
          };
        \addplot[ %
          color=c4,
          mark=*,
          mark size=1.5pt,
          line width=1pt,
        ]
        coordinates {
            (100, 5.152905198776758)
            (1000, 21.978593272171253)
            (10000, 64.3394495412844)
            (274743, 85.2874617737003)
          };
        \addplot[ %
          color=c5,
          mark=*,
          mark size=1.5pt,
          line width=1pt,
        ]
        coordinates {
            (100, 5.17473319183283)
            (1000, 21.918627245822247)
            (10000, 62.72458222474664)
            (274743, 83.58195569639174)
          };
        \addplot[ %
          color=c6,
          mark=*,
          mark size=1.5pt,
          line width=1pt,
        ]
        coordinates {
            (100, 3.5437373159181975)
            (1000, 21.616277254514234)
            (10000, 59.608679814747354)
            (274743, 83.1477337773846)
          };
      \end{groupplot}
      \node[below = 1cm of my plots c1r1.south] {(a)};
      \node[below = 1cm of my plots c2r1.south] {(b)};  
      \node[below = 1cm of my plots c1r2.south] {(c)};
      \node[below = 1cm of my plots c2r2.south] {(d)};      
      \node at ($(my plots c2r1) + (-4.5cm,3.5cm)$){\ref{grouplegend}};
    \end{tikzpicture}
    \caption{Scaling plots of the baseline multilingual mT5 model on test set of \presto{} for: (a) disfluencies, (b) code-mixing, (c) user revisions, and (d) all inclusive (i.e. sampling uniformly from the overall \presto{} training data mixture).}
    \label{fig:data-scaling-main-fig}
  \end{centering}
\end{figure*}
\subsection{Overall Results}
Before diving into focused evaluations on various phenomena of interest, it is instructive to examine the overall model performance on this dataset when trained on 100, 1K, 10K examples, as well as the full the training set.
The results are shown in Fig.~\ref{fig:data-scaling-main-fig}, and they demonstrate a linear increase in model performance as the number of training examples grows exponentially.

Next, we turn our attention to the different linguistic phenomena represented in the dataset and assess how difficult it is for mT5-based models to parse examples with these phenomena. 
Table \ref{tab:0shotphenomena} shows the exact match accuracy for a multilingual model trained on all dataset examples which are not marked for the phenomena discussed earlier.
Each column corresponds to a different test set where all test examples either have user revisions, disfluent utterances, code switching or none of the above.
Unsurprisingly, the results demonstrate that models tend to perform much worse on examples where these phenomena are represented, when the model is not exposed to such phenomena in training.
In contrast, when the model is exposed to a lot of training examples with these phenomena, as in Table \ref{tab:fullshotphenomena}, we do not observe a large gap in performance across the different test sets, except for code-switching where the average performance remains significantly lower despite the high representation of code-switched examples in the training set.

\begin{table*}[h]
\centering
\begin{tabular}{c c c c c c }
\toprule
Language     & No Phenomenon        & User Revisions      &  Disfluency    & Code-Switching   \\
\midrule
German & 81.76 & 23.56 & 64.91 & 56.81 \\ 
English & 85.16 & 22.45 & 63.78 & 67.43 \\ 
Spanish & 81.26 & 19.94 & 61.31 & 62.60 \\ 
Japanese & 83.41 & 22.95 & 54.31 & 71.71 \\ 
Hindi & 76.60 & 26.94 & 54.89 & 58.10 \\ 
French & 84.27 & 15.25 & 67.42 & 64.21 \\ 
\hdashline
Overall & 82.38 &21.51  & 61.26 & 63.16 \\
\bottomrule
\end{tabular}
\caption{Exact match accuracy results (\%) on the test set for the zero-shot multilingual mT5 model (i.e. trained on all examples with no marked phenomena).}
\label{tab:0shotphenomena}
\end{table*}

\subsection{Code Switching}\label{sec:experiments_code_switching}
How do we parse code-switched utterances? To start answering this question, we create a focused test set which only includes the test set examples marked for code-switching.
Specifically, we want to build an intuition for how well an LLM-based model parses code-switched examples in the test set, when fine-tuned on a large number of multilingual examples but only a limited number of code-switching examples.
To this effect, we train several mT5-base models on different training sets which vary in the number of code-switched examples included in training: 0, 5, 25, 125, 625, 3125, and 15,625 examples, and compute their exact match accuracy on the code-switching test set in different languages. 

As shown in Fig.~\ref{fig:data-scaling-main-fig}, when no code-switching examples are used in training, the model's exact match accuracy ranges between 56-72\%, performing best on Japanese and worst on Hindi.  
Surprisingly, even at 0 shot we see a relatively high performance for this phenomenon. We hypothesize that this is due to using a \texttt{LangID} to identify code-switching utterances which may cause code-switched utterances to leak into the training set. 

\subsection{User Revisions}
\label{sec:experiments_revisions}
Similar to the previous section, we ask how many examples with user revisions the model needs to see in training to perform well on user revisions.
We train several mT5-base models on different training sets where the number of examples per user revisions subtype is limited to 0, 5, 25, 125, 625, 3125, and 15,625 examples each (i.e., 5  examples with intent corrections + 5 with argument corrections + 5 with within-turn corrections + 5 with cancellations).
While the training data are dominated by examples that do not have user revisions, the test set used in this section only includes examples with user revisions.

As shown in Fig.~\ref{fig:data-scaling-main-fig}, zero-shot exact match accuracy results range between 18-28\%. Unlike code switching, adding only 25 examples of each type of user revisions notably improves the performance on user revisions. 
After the first few examples, the performance improves linearly as the number of examples with relevant phenomena grows exponentially from 25 to 125 to 625, before it slows down when the number of examples reaches 15K.

\subsection{Disfluencies}
Here, we turn to disfluent utterances and try to estimate how well our models fare on them with various number of disfluent examples in the training set.
Fig.~\ref{fig:data-scaling-main-fig} shows the results for the focused test set with disfluent utterances.
In the zero-shot setup, the performance starts at a higher accuracy range for all languages, compared to use revisions.
As the number of training examples grows exponentially, we observe a linear increase in exact match accuracy for all languages.
\label{sec:experiments_disfluencies}

\subsection{Structured Context}\label{sec:context}
\presto{} provides an opportunity to examine how structured context in the user's virtual environment may be used to improve conversational parsing models.
In this section, we use a focused test set which highlights context by only including contextual examples in the test set (i.e., the first data collection approach discussed in \S\ref{sec:data_collection}).

We linearize and prepend the context tokens to the T5 parser input with \texttt{[SEP]} tokens as shown in the example presented in \ref{para:features}. Initially we simply linearized the entire context and added that as a feature. 
We noticed that naively prepending all the context had too much noise and was not resulting in performance improvements. Hence to represent context in the input sequence, we first identified context notes, contact names and list names which had a trigram similarity greater than 0.6 with the last user utterance and only used the last two turns. We call this `\emph{filtered context representation}' since it filters out parts of the context which are unlikely to provide useful information for the parser.

First, we examine how the performance increases on the focused test set as we include more contextual examples to the training set. 
In particular, we fine-tune the mT5-Base model with filtered context representation on all the non-contextual training examples in addition to 10, 100, 1K or 10K contextual training examples. Table~\ref{tab:contextual} provides average performance for 5 languages with and without context. 
We notice very little improvement with addition of the context features. Our interpretation for this result is that, even though the data contributors were encouraged to use the structured context when conversing with the virtual assistant, most utterances can be understood without referencing the structured context. 
For example, it is easy for the model to parse the utterance ``add apples to my shopping list'', with no knowledge about what lists the user have.(See \S\ref{sec:data_collection} for more details on contextual vs. non-contextual examples.)

\begin{table}[h]
\centering
\resizebox{7cm}{!}{
\begin{tabular}{c c c}
\toprule
Number  & Non-Contextual & Contextual \\
of Examples     & Model    & Model \\
\midrule

10 & 21.23 & 23.94 \\
100 & 40.84 & 41.90 \\
1000 & 64.29 & 64.93 \\
10000 & 80.28 & 80.50 \\
\bottomrule
\end{tabular}
}
\caption{Average performance (exact match accuracy (\%)) on the contextual test set across all the languages for contextual vs non contextual models. It can be seen that in low resource setup contextual models outperform non-contextual model, i.e. modeling context helps.}
\label{tab:contextual}
\end{table}

\subsection{Monolingual vs. Multilingual Models}\label{sec:experiments_monolingual}
\begin{figure}[h]
  \begin{centering}
    \begin{tikzpicture}
      \pgfplotsset{footnotesize,samples=10}
      \begin{groupplot}[
          group style = {group size = 1 by 1, horizontal sep = 3cm},
          width = 7cm,
          height = 7cm]
        \nextgroupplot[
          align = center,
          legend style={at={(0.9,0.05)},anchor=south},
          xmin=100, xmax=50000,
          xmode=log,
          ymin=0, ymax=100,
          axis x line*=top,
          axis y line*=left,
          xlabel={Per Language Training Examples},
          ylabel={Exact Match Accuracy (\%)},
          ytick={0, 10, 20, 30, 40, 50, 60, 70, 80, 90, 100},
          xtick={100, 1000, 10000, 45790},
          xticklabels={100, 1000, 10000, 274743},
          grid style=dashed,
          x label style={at={(axis description cs:0.5,0)},anchor=north},
          y label style={at={(axis description cs:0.1,0.5)},anchor=south},
          xtick pos=bottom,
          ytick pos=left,
          legend style={draw=none},
          legend cell align=right,
          xmode=log,
        ]
        \addplot[ %
          color=c1,
          mark=*,
          mark size=1.5pt,
          line width=1pt,
        ]
        coordinates {
            (100, 7.36)
            (1000, 32.49)
            (10000, 67.66)
            (45790, 79.40)
          };
        \addlegendentry{Monolingual}
        \addplot[ %
          color=c4,
          mark=*,
          mark size=1.5pt,
          line width=1pt,
        ]
        coordinates {
            (100, 14.82)
            (1000, 50.75)
            (10000, 75.44)
            (45790, 81.66)
          };
        \addlegendentry{Multilingual}

      \end{groupplot}
    \end{tikzpicture}
    \caption{Average test set performance of multilingual and monolingual models. In low resource setting, we can see that a multilingual model performs better than a monolingual one.}
    \label{fig:mono-vs-multi}
  \end{centering}
\end{figure}
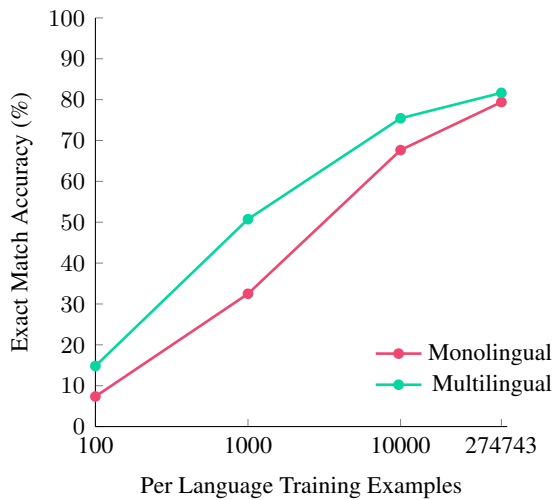

In previous sections, we train a single parser on training examples from all languages in \presto{}, as proposed in \newcite{Ammar2016ManyLO}.
A more traditional approach is to train several monolingual models.
Figure \ref{fig:mono-vs-multi} shows how the monolingual models compare to multilingual models with varying amounts of training data.
Our results show that in lower data regimes there is a clear gap between monolingual and multilingual models, but when using all training instances in the dataset, monolingual and multilingual models converge towards similar performance, in terms of overall accuracy. 

\section{Conclusion}
We introduce \presto{}, a over 550K-examples, six-language dataset for parsing realistic task-oriented dialogs. \presto{} is enriched with a high representation of contextual examples, code switching, user revisions and disfluencies.  Our initial results demonstrate that it is possible for our models to perform well on various linguistic phenomena, but only when the model is exposed to a very large number of training examples with the given phenomenon. We observe that feeding a simple representation of the structured context to the model does not yield large improvements on contextual examples. We also notice that multilingual models perform better than monolingual models in lower data regimes.

\paragraph{Open Questions.}
By introducing a new dataset with unique characteristics, this work deliberately asks more questions than it answers. Some of the obvious modeling questions which this dataset can help answer are: 
How do we train models which perform well on conversational phenomenon like code-switching, user revisions and disfluencies, with a relatively small number of training examples that exhibit these phenomena?
How do we identify examples for which context is critical to understanding the user's utterance?
How can structured context be represented in our models most optimally?
How can we increase cross-lingual supervision in multilingual models to make them more data efficient?
We hope that \presto{} is useful in answering these and other questions about conversational semantic parsing. 

\bibliography{references}

\begin{thebibliography}{18}
\expandafter\ifx\csname natexlab\endcsname\relax\def\natexlab#1{#1}\fi

\bibitem[{Agarwal et~al.(2022)Agarwal, Gupta, Goel, Upadhyay, Joshi, and
  Aravamudhan}]{agarwal2022cst5}
Anmol Agarwal, Jigar Gupta, Rahul Goel, Shyam Upadhyay, Pankaj Joshi, and
  Rengarajan Aravamudhan. 2022.
\newblock \href {https://arxiv.org/abs/2211.07514} {{CST5: Data Augmentation
  for Code-Switched Semantic Parsing}}.
\newblock \emph{arXiv preprint arXiv:2211.07514}.

\bibitem[{Ammar et~al.(2016)Ammar, Mulcaire, Ballesteros, Dyer, and
  Smith}]{Ammar2016ManyLO}
Waleed Ammar, George Mulcaire, Miguel Ballesteros, Chris Dyer, and Noah~A.
  Smith. 2016.
\newblock \href {https://doi.org/10.1162/tacl_a_00109} {{{Many Languages, One
  Parser}}}.
\newblock \emph{Transactions of the Association for Computational Linguistics},
  4:431--444.

\bibitem[{Andreas et~al.(2020)Andreas, Bufe, Burkett, Chen, Clausman, Crawford,
  Crim, DeLoach, Dorner, Eisner, Fang, Guo, Hall, Hayes, Hill, Ho, Iwaszuk,
  Jha, Klein, Krishnamurthy, Lanman, Liang, Lin, Lintsbakh, McGovern,
  Nisnevich, Pauls, Petters, Read, Roth, Roy, Rusak, Short, Slomin, Snyder,
  Striplin, Su, Tellman, Thomson, Vorobev, Witoszko, Wolfe, Wray, Zhang, and
  Zotov}]{smcalflow}
Jacob Andreas, John Bufe, David Burkett, Charles Chen, Josh Clausman, Jean
  Crawford, Kate Crim, Jordan DeLoach, Leah Dorner, Jason Eisner, Hao Fang,
  Alan Guo, David Hall, Kristin Hayes, Kellie Hill, Diana Ho, Wendy Iwaszuk,
  Smriti Jha, Dan Klein, Jayant Krishnamurthy, Theo Lanman, Percy Liang,
  Christopher~H. Lin, Ilya Lintsbakh, Andy McGovern, Aleksandr Nisnevich, Adam
  Pauls, Dmitrij Petters, Brent Read, Dan Roth, Subhro Roy, Jesse Rusak, Beth
  Short, Div Slomin, Ben Snyder, Stephon Striplin, Yu~Su, Zachary Tellman, Sam
  Thomson, Andrei Vorobev, Izabela Witoszko, Jason Wolfe, Abby Wray, Yuchen
  Zhang, and Alexander Zotov. 2020.
\newblock \href {https://doi.org/10.1162/tacl_a_00333} {{{Task-Oriented
  Dialogue as Dataflow Synthesis}}}.
\newblock \emph{Transactions of the Association for Computational Linguistics},
  8:556--571.

\bibitem[{Budzianowski and Vuli{\'c}(2019)}]{Budzianowski2019HelloIG}
Pawe{\l} Budzianowski and Ivan Vuli{\'c}. 2019.
\newblock \href {https://doi.org/10.18653/v1/D19-5602} {{Hello, It{'}s {GPT}-2
  - How Can {I} Help You? Towards the Use of Pretrained Language Models for
  Task-Oriented Dialogue Systems}}.
\newblock In \emph{Proceedings of the 3rd Workshop on Neural Generation and
  Translation}, pages 15--22, Hong Kong. Association for Computational
  Linguistics.

\bibitem[{Budzianowski et~al.(2018)Budzianowski, Wen, Tseng, Casanueva, Ultes,
  Ramadan, and Ga{\v{s}}i{\'c}}]{multiwoz}
Pawe{\l} Budzianowski, Tsung-Hsien Wen, Bo-Hsiang Tseng, I{\~n}igo Casanueva,
  Stefan Ultes, Osman Ramadan, and Milica Ga{\v{s}}i{\'c}. 2018.
\newblock \href {https://doi.org/10.18653/v1/D18-1547} {{{M}ulti{WOZ} - A
  Large-Scale Multi-Domain {W}izard-of-{O}z Dataset for Task-Oriented Dialogue
  Modelling}}.
\newblock In \emph{Proceedings of the 2018 Conference on Empirical Methods in
  Natural Language Processing}, pages 5016--5026, Brussels, Belgium.
  Association for Computational Linguistics.

\bibitem[{Cassell et~al.(2000)Cassell, Bickmore, Campbell, Vilhjalmsson, and
  Yan}]{cassell2000human}
Justine Cassell, Tim Bickmore, Lee Campbell, Hannes Vilhjalmsson, and Hao Yan.
  2000.
\newblock {Human Conversation as a System Framework: Designing Embodied
  Conversational Agents}.
\newblock \emph{Embodied Conversational Agents}, pages 29--63.

\bibitem[{Cheng et~al.(2020)Cheng, Agrawal, Mart{\'\i}nez~Alonso, Bhargava,
  Driesen, Flego, Kaplan, Kartsaklis, Li, Piraviperumal, Williams, Yu,
  {\'O}~S{\'e}aghdha, and Johannsen}]{treedst}
Jianpeng Cheng, Devang Agrawal, H{\'e}ctor Mart{\'\i}nez~Alonso, Shruti
  Bhargava, Joris Driesen, Federico Flego, Dain Kaplan, Dimitri Kartsaklis, Lin
  Li, Dhivya Piraviperumal, Jason~D. Williams, Hong Yu, Diarmuid
  {\'O}~S{\'e}aghdha, and Anders Johannsen. 2020.
\newblock \href {https://doi.org/10.18653/v1/2020.emnlp-main.651}
  {{{Conversational Semantic Parsing for Dialog State Tracking}}}.
\newblock In \emph{Proceedings of the 2020 Conference on Empirical Methods in
  Natural Language Processing (EMNLP)}, pages 8107--8117, Online. Association
  for Computational Linguistics.

\bibitem[{Coucke et~al.(2018)Coucke, Saade, Ball, Bluche, Caulier, Leroy,
  Doumouro, Gisselbrecht, Caltagirone, Lavril et~al.}]{snips}
Alice Coucke, Alaa Saade, Adrien Ball, Th{\'e}odore Bluche, Alexandre Caulier,
  David Leroy, Cl{\'e}ment Doumouro, Thibault Gisselbrecht, Francesco
  Caltagirone, Thibaut Lavril, et~al. 2018.
\newblock \href {https://arxiv.org/abs/1805.10190} {{Snips Voice Platform: an
  embedded Spoken Language Understanding system for private-by-design voice
  interfaces}}.
\newblock \emph{arXiv preprint arXiv:1805.10190}.

\bibitem[{Eric et~al.(2020)Eric, Goel, Paul, Sethi, Agarwal, Gao, Kumar, Goyal,
  Ku, and Hakkani-Tur}]{eric2019multiwoz}
Mihail Eric, Rahul Goel, Shachi Paul, Abhishek Sethi, Sanchit Agarwal, Shuyang
  Gao, Adarsh Kumar, Anuj Goyal, Peter Ku, and Dilek Hakkani-Tur. 2020.
\newblock \href {https://aclanthology.org/2020.lrec-1.53} {{{M}ulti{WOZ} 2.1: A
  Consolidated Multi-Domain Dialogue Dataset with State Corrections and State
  Tracking Baselines}}.
\newblock In \emph{Proceedings of the Twelfth Language Resources and Evaluation
  Conference}, pages 422--428, Marseille, France. European Language Resources
  Association.

\bibitem[{FitzGerald et~al.(2022)FitzGerald, Hench, Peris, Mackie, Rottmann,
  Sanchez, Nash, Urbach, Kakarala, Singh, Ranganath, Crist, Britan, Leeuwis,
  Tur, and Natarajan}]{massive}
Jack FitzGerald, Christopher Hench, Charith Peris, Scott Mackie, Kay Rottmann,
  Ana Sanchez, Aaron Nash, Liam Urbach, Vishesh Kakarala, Richa Singh, Swetha
  Ranganath, Laurie Crist, Misha Britan, Wouter Leeuwis, Gokhan Tur, and Prem
  Natarajan. 2022.
\newblock \href {http://arxiv.org/abs/2204.08582} {{MASSIVE: A 1M-Example
  Multilingual Natural Language Understanding Dataset with 51
  Typologically-Diverse Languages}}.

\bibitem[{Gupta et~al.(2021)Gupta, Xu, Upadhyay, Yang, and
  Faruqui}]{gupta2021disfl}
Aditya Gupta, Jiacheng Xu, Shyam Upadhyay, Diyi Yang, and Manaal Faruqui. 2021.
\newblock \href {https://doi.org/10.18653/v1/2021.findings-acl.293}
  {{{Disfl-{QA}: A Benchmark Dataset for Understanding Disfluencies in Question
  Answering}}}.
\newblock In \emph{Findings of the Association for Computational Linguistics:
  ACL-IJCNLP 2021}, pages 3309--3319, Online. Association for Computational
  Linguistics.

\bibitem[{Gupta et~al.(2018)Gupta, Shah, Mohit, Kumar, and Lewis}]{top}
Sonal Gupta, Rushin Shah, Mrinal Mohit, Anuj Kumar, and Mike Lewis. 2018.
\newblock \href {https://doi.org/10.18653/v1/D18-1300} {{{Semantic Parsing for
  Task Oriented Dialog using Hierarchical Representations}}}.
\newblock In \emph{Proceedings of the 2018 Conference on Empirical Methods in
  Natural Language Processing}, pages 2787--2792, Brussels, Belgium.
  Association for Computational Linguistics.

\bibitem[{Kim et~al.(2022)Kim, Liu, Jin, Papangelis, Hedayatnia,
  Gopalakrishnan, and Hakkani-Tür}]{dstc10}
Seokhwan Kim, Yang Liu, Di~Jin, Alexandros Papangelis, Behnam Hedayatnia,
  Karthik Gopalakrishnan, and Dilek Hakkani-Tür. 2022.
\newblock \href
  {https://www.amazon.science/publications/knowledge-grounded-task-oriented-dialogue-modeling-on-spoken-conversations-track-at-dstc10}
  {{Knowledge-grounded task-oriented dialogue modeling on spoken conversations
  track at DSTC10}}.
\newblock In \emph{AAAI 2022 Workshop on Dialog System Technology Challenge}.

\bibitem[{Li et~al.(2021)Li, Arora, Chen, Gupta, Gupta, and Mehdad}]{mtop}
Haoran Li, Abhinav Arora, Shuohui Chen, Anchit Gupta, Sonal Gupta, and Yashar
  Mehdad. 2021.
\newblock \href {https://doi.org/10.18653/v1/2021.eacl-main.257} {{{MTOP}: A
  Comprehensive Multilingual Task-Oriented Semantic Parsing Benchmark}}.
\newblock In \emph{Proceedings of the 16th Conference of the European Chapter
  of the Association for Computational Linguistics: Main Volume}, pages
  2950--2962, Online. Association for Computational Linguistics.

\bibitem[{Pasupat et~al.(2021)Pasupat, Zhang, and
  Guu}]{Pasupat2021ControllableSP}
Panupong Pasupat, Yuan Zhang, and Kelvin Guu. 2021.
\newblock \href {https://doi.org/10.18653/v1/2021.emnlp-main.607}
  {{{Controllable Semantic Parsing via Retrieval Augmentation}}}.
\newblock In \emph{Proceedings of the 2021 Conference on Empirical Methods in
  Natural Language Processing}, pages 7683--7698, Online and Punta Cana,
  Dominican Republic. Association for Computational Linguistics.

\bibitem[{Roberts et~al.(2022)Roberts, Chung, Levskaya, Mishra, Bradbury,
  Andor, Narang, Lester, Gaffney, Mohiuddin, Hawthorne, Lewkowycz, Salcianu,
  van Zee, Austin, Goodman, Soares, Hu, Tsvyashchenko, Chowdhery, Bastings,
  Bulian, Garcia, Ni, Chen, Kenealy, Clark, Lee, Garrette, Lee-Thorp, Raffel,
  Shazeer, Ritter, Bosma, Passos, Maitin-Shepard, Fiedel, Omernick, Saeta,
  Sepassi, Spiridonov, Newlan, and Gesmundo}]{Roberts2022ScalingUM}
Adam Roberts, Hyung~Won Chung, Anselm Levskaya, Gaurav Mishra, James Bradbury,
  Daniel Andor, Sharan Narang, Brian Lester, Colin Gaffney, Afroz Mohiuddin,
  Curtis Hawthorne, Aitor Lewkowycz, Alex Salcianu, Marc van Zee, Jacob Austin,
  Sebastian Goodman, Livio~Baldini Soares, Haitang Hu, Sasha Tsvyashchenko,
  Aakanksha Chowdhery, Jasmijn Bastings, Jannis Bulian, Xavier Garcia, Jianmo
  Ni, Andrew Chen, Kathleen Kenealy, Jonathan~H. Clark, Stephan Lee, Dan
  Garrette, James Lee-Thorp, Colin Raffel, Noam Shazeer, Marvin Ritter, Maarten
  Bosma, Alexandre Passos, Jeremy Maitin-Shepard, Noah Fiedel, Mark Omernick,
  Brennan Saeta, Ryan Sepassi, Alexander Spiridonov, Joshua Newlan, and Andrea
  Gesmundo. 2022.
\newblock \href {https://arxiv.org/abs/2203.17189} {{Scaling Up Models and Data
  with $\texttt{t5x}$ and $\texttt{seqio}$}}.
\newblock \emph{arXiv preprint arXiv:2203.17189}.

\bibitem[{Xu et~al.(2020)Xu, Haider, and Mansour}]{multiatis}
Weijia Xu, Batool Haider, and Saab Mansour. 2020.
\newblock \href {https://doi.org/10.18653/v1/2020.emnlp-main.410} {{End-to-End
  Slot Alignment and Recognition for Cross-Lingual {NLU}}}.
\newblock In \emph{Proceedings of the 2020 Conference on Empirical Methods in
  Natural Language Processing (EMNLP)}, pages 5052--5063, Online. Association
  for Computational Linguistics.

\bibitem[{Xue et~al.(2021)Xue, Constant, Roberts, Kale, Al-Rfou, Siddhant,
  Barua, and Raffel}]{Xue2020mT5AM}
Linting Xue, Noah Constant, Adam Roberts, Mihir Kale, Rami Al-Rfou, Aditya
  Siddhant, Aditya Barua, and Colin Raffel. 2021.
\newblock \href {https://doi.org/10.18653/v1/2021.naacl-main.41} {{{m{T}5: A
  Massively Multilingual Pre-trained Text-to-Text Transformer}}}.
\newblock In \emph{Proceedings of the 2021 Conference of the North American
  Chapter of the Association for Computational Linguistics: Human Language
  Technologies}, pages 483--498, Online. Association for Computational
  Linguistics.

\end{thebibliography}

\section{Appendix}
We present the results on the overall dataset as well as the model scaling results in this section. Table ~\ref{tab:fullshotphenomena} presents results on the testsets if we use all the available training data.

\begin{figure*}[t]
  \begin{centering}
    \begin{tikzpicture}
      \pgfplotsset{footnotesize,samples=10}
      \begin{groupplot}[
          group style = {group size = 2 by 2, horizontal sep = 3cm, vertical sep = 2cm},
          width = 6.5cm,
          height = 6.5cm]
        \nextgroupplot[
          align = left,
          xmin=0.58, xmax=13,
          xmode=log,
          ymin=55, ymax=70,
          axis x line*=top,
          axis y line*=left,
          xlabel={Overall data},
          ylabel={Exact Match Accuracy (\%)},
          ytick={55, 60, 65, 70},
          xtick={0.58, 1.3, 3.7, 13},
          xticklabels={0.58, 1.3, 3.7, 13},
          grid style=dashed,
          x label style={at={(axis description cs:0.5,0)},anchor=north},
          y label style={at={(axis description cs:0.1,0.5)},anchor=south},
          xtick pos=bottom,
          ytick pos=left,
          legend style = { /tikz/every even column/.append style={column sep=0.5cm}, legend columns = -1, legend to name = grouplegend2,},  
          xmode=log,
        ]
        \addplot[ %
          color=c1,
          mark=*,
          mark size=1.5pt,
          line width=1pt,
        ]
        coordinates {
(0.58,59.276927795004596)
(1.2,61.00112447609637)
(3.7,62.35731079837803)
(13,64.70167308413126)
          };
        \addlegendentry{de-DE}
        \addplot[ %
          color=c2,
          mark=*,
          mark size=1.5pt,
          line width=1pt,
        ]
        coordinates {
(0.58,58.363916183260336)
(1.2,61.88390355550294)
(3.7,62.92963971429699)
(13,66.0352787972061)
          };
        \addlegendentry{hi-IN}
        \addplot[ %
          color=c3,
          mark=*,
          mark size=1.5pt,
          line width=1pt,
        ]
        coordinates {
(0.58,56.670122944748925)
(1.2,58.74092727003407)
(3.7,60.56880462153755)
(13,63.20545104428974)

          };
        \addlegendentry{es-ES}
        \addplot[ %
          color=c4,
          mark=*,
          mark size=1.5pt,
          line width=1pt,
        ]
        coordinates {
 (0.58,61.31192660550459)
(1.2,62.82874617737003)
(3.7,64.6085626911315)
(13,67.53516819571865)
          };
        \addlegendentry{en-US}
        \addplot[ %
          color=c5,
          mark=*,
          mark size=1.5pt,
          line width=1pt,
        ]
        coordinates {
(0.58,59.82182894382828)
(1.2,62.150608352515626)
(3.7,63.47792293204987)
(13,65.14603449822128)
          };
        \addlegendentry{fr-FR}       
        \addplot[ %
          color=c6,
          mark=*,
          mark size=1.5pt,
          line width=1pt,
        ]
        coordinates {
     (0.58,59.16896497892491)
(1.2,59.652911484623)
(3.7,61.86449497840454)
(13,63.48545558619973)
          };
        \addlegendentry{ja-JP}
        \nextgroupplot[
          align = center,
          xmin=0.58, xmax=13,
          xmode=log,
          ymin=10, ymax=35,
          axis x line*=top,
          axis y line*=left,
          xlabel={User revisions},
          ylabel={Exact Match Accuracy (\%)},
          ytick={10, 15, 20, 25, 30, 35},
          xtick={0.58, 1.3, 3.7, 13},
          xticklabels={0.58, 1.3, 3.7, 13},
          grid style=dashed,
          x label style={at={(axis description cs:0.5,0)},anchor=north},
          y label style={at={(axis description cs:0.1,0.5)},anchor=south},
          xtick pos=bottom,
          ytick pos=left,
          xmode=log,
        ]
        \addplot[ %
          color=c1,
          mark=*,
          mark size=1.5pt,
          line width=1pt,
        ]
        coordinates {
(0.58,23.564254442441413)
(1.2,20.26783414885398)
(3.7,22.418233324748908)
(13,25.032191604429567)
          };
        \addplot[ %
          color=c2,
          mark=*,
          mark size=1.5pt,
          line width=1pt,
        ]
        coordinates {
 (0.58,26.936872309899567)
(1.2,28.281922525107607)
(3.7,28.676470588235293)
(13,34.164275466284074)

          };
        \addplot[ %
          color=c3,
          mark=*,
          mark size=1.5pt,
          line width=1pt,
        ]
        coordinates {
   (0.58,19.940415040065748)
(1.2,17.454283953153894)
(3.7,19.570577357715223)
(13,24.470926648859667)
          };
        \addplot[ %
          color=c4,
          mark=*,
          mark size=1.5pt,
          line width=1pt,
        ]
        coordinates {
(0.58,22.448141302115424)
(1.2,19.95276237420415)
(3.7,21.94495789689875)
(13,28.86629698089957)
          };
        \addplot[ %
          color=c5,
          mark=*,
          mark size=1.5pt,
          line width=1pt,
        ]
        coordinates {
(0.58,15.25)
(1.2,13.318181818181818)
(3.7,14.670454545454545)
(13,17.079545454545457)

          };
        \addplot[ %
          color=c6,
          mark=*,
          mark size=1.5pt,
          line width=1pt,
        ]
        coordinates {
(0.58,22.946859903381643)
(1.2,17.352657004830917)
(3.7,19.246376811594203)
(13,23.053140096618357)

          };
        \nextgroupplot[
          align = center,
          legend style={at={(0.9,0.05)},anchor=south},
          xmin=.58, xmax=13,
          xmode=log,
          ymin=50, ymax=80,
          axis x line*=top,
          axis y line*=left,
          xlabel={Code-mixing},
          ylabel={Exact Match Accuracy (\%)},
          ytick={50, 55, 60, 65, 70, 75, 80},
             xtick={0.58, 1.3, 3.7, 13},
          xticklabels={0.58, 1.3, 3.7, 13},
          grid style=dashed,
          x label style={at={(axis description cs:0.5,0)},anchor=north},
          y label style={at={(axis description cs:0.1,0.5)},anchor=south},
          xtick pos=bottom,
          ytick pos=left,
          legend style={draw=none},
          legend cell align=right,
          xmode=log,
        ]
        \addplot[ %
          color=c1,
          mark=*,
          mark size=1.5pt,
          line width=1pt,
        ]
        coordinates {
   (0.58,56.81139444061567)
(1.2,61.40592694693316)
(3.7,61.8653801975649)
(13,62.87617734895474)
          };
        \addplot[ %
          color=c2,
          mark=*,
          mark size=1.5pt,
          line width=1pt,
        ]
        coordinates {
(0.58,58.097863542384566)
(1.2,62.68090971743625)
(3.7,64.57615437629222)
(13,66.14403859407305)
          };
        \addplot[ %
          color=c3,
          mark=*,
          mark size=1.5pt,
          line width=1pt,
        ]
        coordinates {
  (0.58,62.60180995475113)
(1.2,67.14932126696831)
(3.7,67.23981900452489)
(13,69.16289592760181)

          };
        \addplot[ %
          color=c4,
          mark=*,
          mark size=1.5pt,
          line width=1pt,
        ]
        coordinates {
(0.58,67.43185078909613)
(1.2,74.22285987565758)
(3.7,74.98804399808704)
(13,75.5141080822573)

          };
        \addplot[ %
          color=c5,
          mark=*,
          mark size=1.5pt,
          line width=1pt,
        ]
        coordinates {
(0.58,64.2072902942468)
(1.2,69.85068072024593)
(3.7,69.74088713219147)
(13,70.68511198945981)
          };
        \addplot[ %
          color=c6,
          mark=*,
          mark size=1.5pt,
          line width=1pt,
        ]
        coordinates {
(0.58,71.7065868263473)
(1.2,75.50523952095809)
(3.7,76.08532934131736)
(13,77.71332335329342)
          };
        \nextgroupplot[
          align = center,
          legend style={at={(0.9,0.05)},anchor=south},
          xmin=0.58, xmax=13,
          xmode=log,
          ymin=50, ymax=80,
          axis x line*=top,
          axis y line*=left,
          xlabel={Disfluencies},
          ylabel={Exact Match Accuracy (\%)},
          ytick={50, 55, 60, 65, 70, 75, 80},
          xtick={0.58, 1.3, 3.7, 13},
          xticklabels={0.58, 1.3, 3.7, 13},
          grid style=dashed,
          x label style={at={(axis description cs:0.5,0)},anchor=north},
          y label style={at={(axis description cs:0.1,0.5)},anchor=south},
          xtick pos=bottom,
          ytick pos=left,
          legend style={draw=none},
          legend cell align=right,
          xmode=log,
        ]
        \addplot[ %
          color=c1,
          mark=*,
          mark size=1.5pt,
          line width=1pt,
        ]
        coordinates {
 (0.58,64.91347132677299)
(1.2,68.49338310145912)
(3.7,70.68204954190702)
(13,76.06040040719375)
          };
        \addplot[ %
          color=c2,
          mark=*,
          mark size=1.5pt,
          line width=1pt,
        ]
        coordinates {
(0.58,54.892667967661005)
(1.2,59.54836911067745)
(3.7,60.27320880959019)
(13,66.12768330080847)

          };
        \addplot[ %
          color=c3,
          mark=*,
          mark size=1.5pt,
          line width=1pt,
        ]
        coordinates {
     (0.58,61.3082627118644)
(1.2,66.80349576271186)
(3.7,70.70974576271186)
(13,73.79502118644068)
          };
        \addplot[ %
          color=c4,
          mark=*,
          mark size=1.5pt,
          line width=1pt,
        ]
        coordinates {
         (0.58,63.77914507772021)
(1.2,67.53562176165802)
(3.7,72.31217616580311)
(13,74.49805699481865)
          };
        \addplot[ %
          color=c5,
          mark=*,
          mark size=1.5pt,
          line width=1pt,
        ]
        coordinates {
    (0.58,67.41652518392756)
(1.2,72.2693831352575)
(3.7,75.96208262591963)
(13,78.96151669496322)
          };
        \addplot[ %
          color=c6,
          mark=*,
          mark size=1.5pt,
          line width=1pt,
        ]
        coordinates {
      (0.58,54.30519014589811)
(1.2,56.17077254245396)
(3.7,62.4970102846209)
(13,62.65247548433389)
          };
      \end{groupplot}
      \node at ($(group c2r1) + (-4.5cm,3.5cm)$){\ref{grouplegend2}};
    \end{tikzpicture}
    \caption{Scaling plots for increasing the model capacity. This scaling study was done using 125 shot training set.  As expected larger models generally perform better. }
    \label{fig:model-scaling-main-fig}
  \end{centering}
\end{figure*}
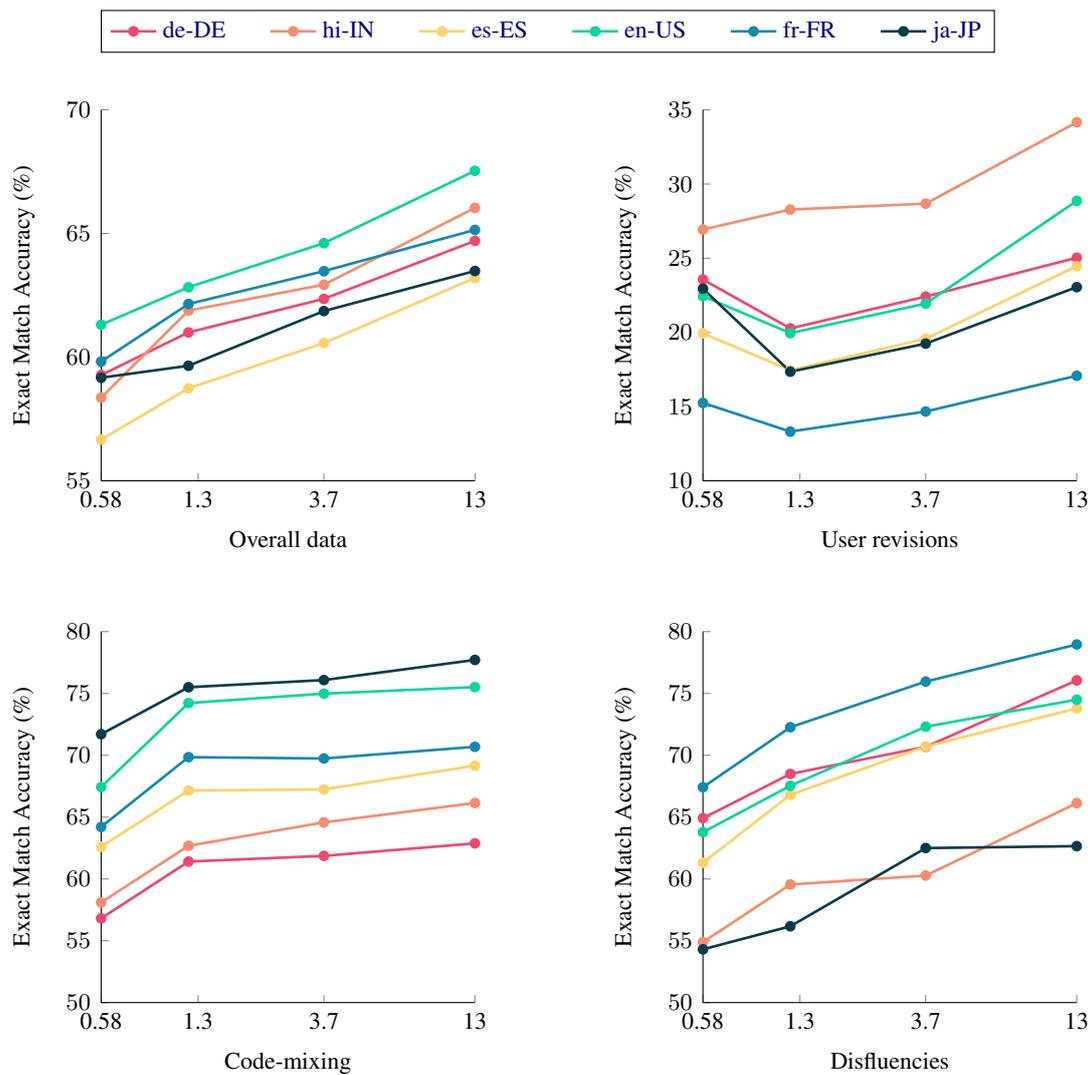

\subsection{Model Scaling}\label{sec:experiments_model_scaling}

The experiments presented so far use mT5-Base since it incurs lower cost than larger models and provides reasonable performance on many NLP tasks.
In this section, we examine larger model sizes impact the results on this dataset.
As a case study, we focus on the 125-shot models for code switching, disfluencies and user revisions.
Fig.~\ref{fig:model-scaling-main-fig} demonstrates the impact of increased model size on the overall test results.
As expected, the larger models consistently give better results but are still lower than the best results we were able to achieve by using the full training set with mT5-Base (discussed earlier in Fig.~\ref{fig:data-scaling-main-fig}).
Fig.~\ref{fig:model-scaling-main-fig} demonstrates a similar pattern for the focused evaluation on user revisions, and we see a similar pattern for other phenomena.
These results confirm that mT5-Base strikes a good balance between model size and performance, and we recommend using mT5-Base for extensions of this work.

\begin{table*}[] 
\centering
\begin{tabular}{lcccccc}
\toprule
\textbf{Intent}           &  \textbf{German}  & \textbf{English}  & \textbf{Spanish}  & \textbf{French } & \textbf{Hindi}  & \textbf{Japanese}   \\\midrule
\textbf{Add\_contact}           & 2779  & 2931  & 3382  & 3148  & 2122  & 3527   \\
\textbf{Add\_item\_to\_list}    & 944   & 2959  & 2913  & 986   & 1958  & 2503   \\
\textbf{BuyEventTickets}        & 2550  & 2463  & 2883  & 2842  & 1697  & 3049   \\
\textbf{Cancel}                 & 1479  & 3029  & 2061  & 1775  & 2318  & 2525   \\
\textbf{Cancel\_ride}           & 2252  & 2353  & 3239  & 3047  & 2053  & 3098   \\
\textbf{Check\_order\_status}   & 2700  & 2877  & 3448  & 3455  & 2136  & 3934   \\
\textbf{Create\_list}           & 844   & 594   & 745   & 1063  & 778   & 1023   \\
\textbf{Create\_note}           & 1566  & 1700  & 1903  & 1819  & 1568  & 1998   \\
\textbf{Find\_parking}          & 4562  & 1210  & 4387  & 2345  & 3117  & 2692   \\
\textbf{GetGenericBusinessType} & 3130  & 2613  & 2782  & 2904  & 1676  & 3139   \\
\textbf{Get\_bill}              & 2249  & 2466  & 2922  & 2582  & 1857  & 3275   \\
\textbf{Get\_health\_stats}     & 2477  & 2512  & 2827  & 2847  & 1982  & 2879   \\
\textbf{Get\_list}              & 701   & 705   & 698   & 676   & 856   & 761    \\
\textbf{Get\_message\_content}  & 2102  & 2142  & 2371  & 2343  & 1320  & 2879   \\
\textbf{Get\_note}              & 1737  & 1905  & 2022  & 2277  & 1978  & 2138   \\
\textbf{Get\_product}           & 2341  & 2076  & 2483  & 2438  & 1574  & 3344   \\
\textbf{Get\_security\_price}   & 2175  & 2333  & 2707  & 2425  & 1752  & 3245   \\
\textbf{Initiate\_call}         & 1822  & 4292  & 2239  & 1965  & 1807  & 2250   \\
\textbf{Log\_exercise}          & 2602  & 2708  & 2050  & 2663  & 1962  & 3049   \\
\textbf{Log\_nutrition}         & 2219  & 2459  & 2667  & 2605  & 1786  & 3204   \\
\textbf{Open\_app}              & 2488  & 2497  & 3367  & 3076  & 2069  & 3643   \\
\textbf{Order\_menu\_item}      & 224   & 860   & 659   & 736   & 155   & 740    \\
\textbf{Order\_ride}            & 2175  & 2342  & 2857  & 2992  & 1809  & 2993   \\
\textbf{Other}                  & 8457  & 17499 & 5650  & 10937 & 9163  & 13737  \\
\textbf{Pause\_exercise}        & 2821  & 2995  & 3109  & 3149  & 1830  & 3495   \\
\textbf{Pay\_bill}              & 2280  & 2304  & 2837  & 2823  & 1834  & 3048   \\
\textbf{Play\_game}             & 2640  & 2667  & 2889  & 3048  & 1941  & 3074   \\
\textbf{Post\_message}          & 2597  & 2224  & 3065  & 3278  & 2049  & 3529   \\
\textbf{Record\_video}          & 2609  & 2295  & 3317  & 3369  & 1934  & 3864   \\
\textbf{Resume\_exercise}       & 2367  & 2060  & 3136  & 2929  & 1991  & 2896   \\
\textbf{Send\_digital\_object}  & 2307  & 3127  & 2804  & 2652  & 2016  & 2905   \\
\textbf{Start\_exercise}        & 2719  & 3030  & 3272  & 3182  & 2031  & 3311   \\
\textbf{Stop\_exercise}         & 2719  & 2814  & 3207  & 3127  & 2078  & 3432   \\
\textbf{Take\_photo}            & 3950  & 2630  & 5266  & 4367  & 4910  & 4349   \\ \hline
\textbf{Overall}                  & 83584 & 95671 & 96164 & 95870 & 72107 & 109528 \\
\bottomrule
\end{tabular}
\caption{Intent distribution by language for \presto{}.}
\label{tab:intent_distribution}
\end{table*}

\begin{table*}[]
\centering
\begin{tabular}{c c c c c }
\toprule
Language     & No Phenomenon        & User Revisions      &  Disfluency    & Code-Switching   \\
\midrule
German & 82.41 & 81.33 & 84.19 & 72.50 \\ 
English & 86.18 & 85.36 & 86.20 & 75.99 \\ 
Spanish & 82.95 & 82.26 & 82.81 & 72.22 \\ 
Japanese & 84.91 & 82.22 & 80.46 & 84.41 \\ 
Hindi & 78.30 & 73.24 & 75.49 & 73.24 \\ 
French & 85.42 & 82.99 & 85.82 & 75.98 \\ 
\hdashline
Overall & 83.66     & 81.85    & 82.92       & 75.88  \\
\bottomrule
\end{tabular}
\caption{Exact match accuracy results (\%) for the multilingual mT5 model trained on the full training set of \presto{}.}
\label{tab:fullshotphenomena}
\end{table*}

\clearpage

\end{document}